\documentclass[runningheads]{llncs}

\usepackage{eccv}

\usepackage{eccvabbrv}

\usepackage{graphicx}
\usepackage{booktabs}

\usepackage[accsupp]{axessibility}  

\usepackage[hidelinks]{hyperref}

\usepackage{orcidlink}

\begin{document}

\title{Multi-scale Object-Aware Gaze Estimation via Geometric Reasoning} 


\author{Jiajie Mi\textsuperscript{*}\orcidlink{0009-0008-7835-9300} \and
Xinyu Liu\textsuperscript{*}\orcidlink{0009-0004-4882-7987} \and
Mengke Song\orcidlink{0000-0001-9618-0656} \and
Chenglizhao Chen\textsuperscript{\textdagger}\orcidlink{0000-0001-9982-5667}}

\authorrunning{J.~Mi et al.}

\institute{Qingdao Institute of Software \& School of Computer Science and Technology, China University of Petroleum (East China), China\\
Shandong Key Laboratory of Intelligent Oil \& Gas Industrial Software, China\\
\email{cclz123@163.com}}

\maketitle

\let\eccvsavedfn\thefootnote
\let\thefootnote\relax
\footnotetext{\textsuperscript{*}~Equal contribution.\quad\textsuperscript{\textdagger}~Corresponding author.}
\let\thefootnote\eccvsavedfn

\begin{abstract}
    
Gaze target estimation aims to predict the semantic object an observer fixates upon within an image, a task deeply rooted in the object-oriented nature of human gaze.
Observers tend to select a specific semantic entity as the attentional target, rather than responding randomly across arbitrary regions of the image.
However, existing methods typically model this task as a direct mapping from global features to gaze heatmaps, essentially treating it as a pixel-level regression problem.
This approach fails to explicitly represent the gazed object as a distinct entity, making it difficult to produce stable and semantically consistent predictions in complex scenes.
To address this, we propose a two-stage gaze estimation framework guided by object semantics, reformulating gaze target estimation as a hierarchical reasoning process.
Our method incorporates object-level representations during feature encoding to align image features with discrete semantic entities, then introduces multi-scale feature fusion and geometric constraints from head pose and gaze direction for fine-grained localization and object-level discrimination.
Extensive experiments on GazeFollow, VideoAttentionTarget, ChildPlay, and GOO-Real demonstrate that our method achieves AUC of 0.961, 0.948, 0.987, and 0.977 respectively, delivering strong performance across all benchmarks while maintaining a compact parameter size of 7.1M.

\keywords{Gaze Target Estimation \and Object-Aware Fusion \and Multi-Scale Feature Fusion \and Geometric Constraints}
\end{abstract}

\section{Introduction}
\label{sec:intro}

\begin{figure}[!t]
      \centering
      \includegraphics[width=\textwidth]{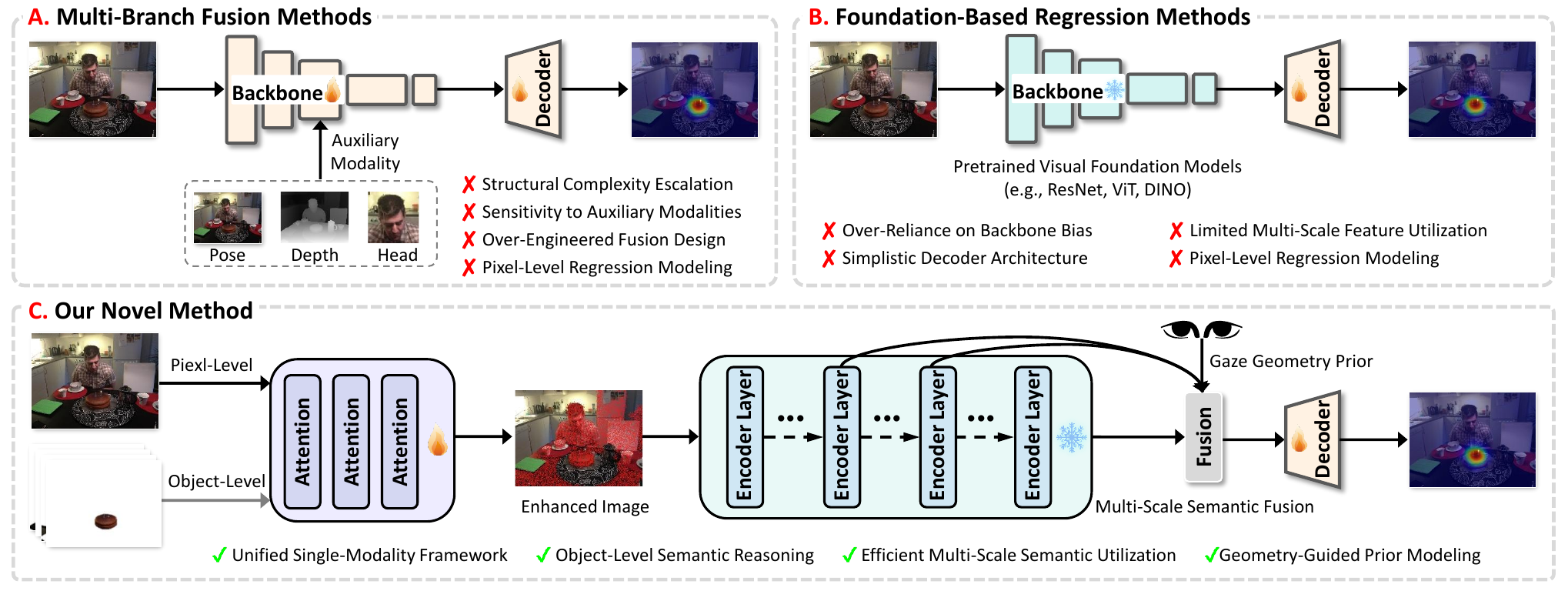}
      \caption{Comparison of existing gaze estimation paradigms and the proposed method.
      	(A) Multi-branch fusion methods.
      	(B) Foundation-based regression methods.
      	(C) Our method reformulates gaze target estimation as hierarchical reasoning, enabling object-level modeling and multi-scale localization in a unified single-modality framework.}
      \label{fig:motivation}
\end{figure}

Gaze target estimation aims to predict the location of the object that an observer is looking at in a scene. It is a fundamental task for understanding human visual attention and social interaction\cite{emery2000eyes,admoni2017social,fan2019understanding,fan2018inferring}. In autonomous driving\cite{deng2016does,zhao2023improving,liu2019gaze}, a driver's gaze target reflects attention state and awareness of potential risks. In human-computer interaction\cite{majaranta2019eye,admoni2017social,madhusanka2022biofeedback,beyan2023co} and social robotics\cite{zhang2023human,arreghini2024predicting,holman2021watch,prada2023gaze}, accurate understanding of gaze targets is essential for joint attention\cite{frischen2007gaze,moore2014joint,thoermer2001preverbal} and natural interaction\cite{fischer2018rt,canigueral2019role}. Collectively, these scenarios establish gaze target estimation as a fundamental research problem in computer vision. 

Although significant progress has been made in recent years, existing methods commonly formulate gaze target estimation as a direct regression from image features to spatial heatmaps. Current approaches can be broadly divided into two categories. As shown in Fig.~\ref{fig:motivation}(A), one line of work introduces auxiliary modalities such as depth\cite{bao2022escnet,fang2021dual,guan2019enhanced,jin2022depth,miao2023patch}, human pose\cite{bao2022escnet,yang2024gaze}, or three-dimensional head information\cite{bao2022escnet,miao2025multi} and builds multi-branch networks\cite{wang2026vl,bao2022escnet,chong2020detecting,fang2021dual,gupta2022modular,horanyi2023they,jin2022depth} to enhance spatial understanding. However, as more modalities are added, model complexity increases continuously. Prediction performance becomes highly dependent on the accuracy of upstream auxiliary models, which often leads to limited generalization across different scenes or real world environments. As shown in Fig.~\ref{fig:motivation}(B), another line of work adopts pretrained visual foundation models as unified feature extractors. This design improves parameter efficiency and simplifies network architecture, but prediction is usually performed using only the final layer features through shallow decoding. As a result, important spatial structure information contained in hierarchical representations is largely ignored. More importantly, both categories share the same underlying modeling assumption, where gaze prediction is treated as a pixel-level regression problem.

This modeling paradigm overlooks a fundamental property of human visual attention that human gaze is naturally object oriented. In real visual behavior, an observer first selects a specific semantic object as the focus of attention rather than responding randomly to arbitrary locations in the image. In other words, the gaze process consists of two consecutive stages. The observer first determines which object is being attended, and then forms a precise gaze location within that object region. When models lack explicit object-level decision modeling, predictions often become spatially scattered or semantically unstable in complex scenes where multiple candidate targets compete for attention. This observation suggests that the main limitation of existing methods does not arise from insufficient feature representation, but from the absence of a structured modeling mechanism for object selection in task formulation.

Based on the above observations, this work reformulates gaze target estimation as a hierarchical reasoning process. The model first constructs a semantic hypothesis space of potential gaze objects within the scene and then incorporates geometric constraints of the observer to determine which regions are visually reachable. Precise localization is subsequently performed within candidate regions that satisfy both semantic and geometric consistency. This perspective transforms gaze prediction from single-step pixel regression into a reasoning problem composed of object-level discrimination and region-level localization, making the modeling process more consistent with human visual behavior.

Following this formulation, we propose a multi-scale object-aware framework for gaze target estimation. As illustrated in Fig.~\ref{fig:motivation}(C), the framework adopts a pretrained visual foundation model as a unified feature extractor to maintain parameter efficiency while enabling structured reasoning through three key mechanisms. First, object-level semantic representations are introduced during feature encoding to establish explicit associations between scene features and discrete semantic entities. Second, multi-scale feature fusion is applied to fully utilize spatial and semantic information across different hierarchical layers of the foundation model. Third, a field-of-view geometric constraint derived from gaze direction is introduced to provide a physiologically plausible spatial prior. These components together form a complete reasoning pipeline from candidate object identification to precise gaze localization in complex real-world scenes.

We evaluate our method on GazeFollow~\cite{recasens2015they}, VideoAttentionTarget~\cite{chong2020detecting}, ChildPlay~\cite{tafasca2023childplay}, and GOO-Real~\cite{tomas2021goo}. 
Our main contributions are as follows:
\begin{itemize}
\item We revisit gaze target estimation from a cognitive modeling perspective and reformulate it from pixel-level regression into a hierarchical reasoning problem that jointly performs object selection and spatial localization.

\item We develop a unified single-modality framework that realizes the proposed hierarchical reasoning formulation through object-level semantic modeling, multi-scale feature representation, and gaze direction guidance.

\item We demonstrate the proposed framework achieves strong and consistent performance on four benchmark datasets, with AUC scores of 0.961, 0.948, 0.987, and 0.977 on GazeFollow, VideoAttentionTarget, ChildPlay, and GOO-Real respectively, while maintaining a compact 7.1M-parameter model.
\end{itemize}

\section{Related Work}

\textbf{Pixel-Level Gaze Estimation.}
Early gaze target estimation methods formulate gaze prediction as a direct mapping from visual features to spatial heatmaps. Representative approaches adopt dual-branch architectures\cite{chen2021gaze, chong2018connecting, chong2020detecting, lian2018believe, recasens2015they, saran2018human} to jointly encode head appearance and scene context in order to infer gaze locations. Subsequent works enhance this paradigm by incorporating attention mechanisms\cite{song2024vitgaze,tu2023gaze,tafasca2024sharingan}, global context modeling\cite{wang2022gatector,tafasca2024sharingan,tu2023gaze}, or temporal aggregation\cite{gupta2024mtgs,miao2024diffusion,miao2023patch,yu2026modality,yu2026anisotropic} to improve feature interaction between observers and surrounding regions. Despite architectural improvements, these methods fundamentally treat gaze estimation as a pixel-level regression problem, where predictions are generated directly from global feature responses without explicitly modeling the attended object. Such formulations often lead to spatially diffused predictions when multiple candidate objects coexist within complex scenes.

\noindent\textbf{Object-Level Gaze Representation.}
Recognizing the object-oriented nature of human attention, several studies attempt to introduce object information into gaze estimation. Object-aware approaches\cite{nieva2025towards,chen2026controlling,jin2025gatector+} leverage detected regions or semantic cues to guide attention modeling and improve interpretability. These methods demonstrate that incorporating object-level cues can reduce ambiguity caused by cluttered backgrounds or competing targets. However, object information is typically used as auxiliary guidance rather than being integrated into the core decision process. As a result, gaze prediction remains largely dependent on feature aggregation or regression-based decoding, lacking an explicit mechanism that models gaze estimation as object selection followed by spatial localization.

\noindent\textbf{Foundation Model Based Gaze Estimation.}
Recent advances in visual foundation models\cite{song2024vitgaze,ryan2025gaze,lan2026fgi,miao2024diffusion,yang2025gazellm} enable gaze estimation frameworks to benefit from strong pretrained representations. Methods built upon large-scale encoders improve parameter efficiency and generalization capability by leveraging frozen backbones. Nevertheless, most approaches rely primarily on final-layer features for prediction, underutilizing hierarchical representations available within deep networks. More importantly, these methods continue to adopt regression-based formulations, where gaze prediction is inferred from feature responses rather than structured reasoning over candidate objects. Consequently, the object-oriented nature of gaze behavior remains insufficiently modeled.

\section{Method}
\label{sec:method}

\subsection{Hierarchical Reasoning Overview}

Gaze target estimation is reformulated as a hierarchical reasoning process that progressively narrows the search space from global scenes to precise fixation locations. Rather than directly regressing gaze heatmaps from global image features, the proposed framework models gaze prediction as structured inference over semantically meaningful and geometrically feasible regions. The model first establishes object-level semantic hypotheses by identifying candidate entities that may attract visual attention, while observer-dependent geometric cues derived from head position and gaze direction constrain attention to spatially plausible regions consistent with human visual behavior and natural attention patterns.

\begin{figure}[!t]
    \centering
    \includegraphics[width=\textwidth]{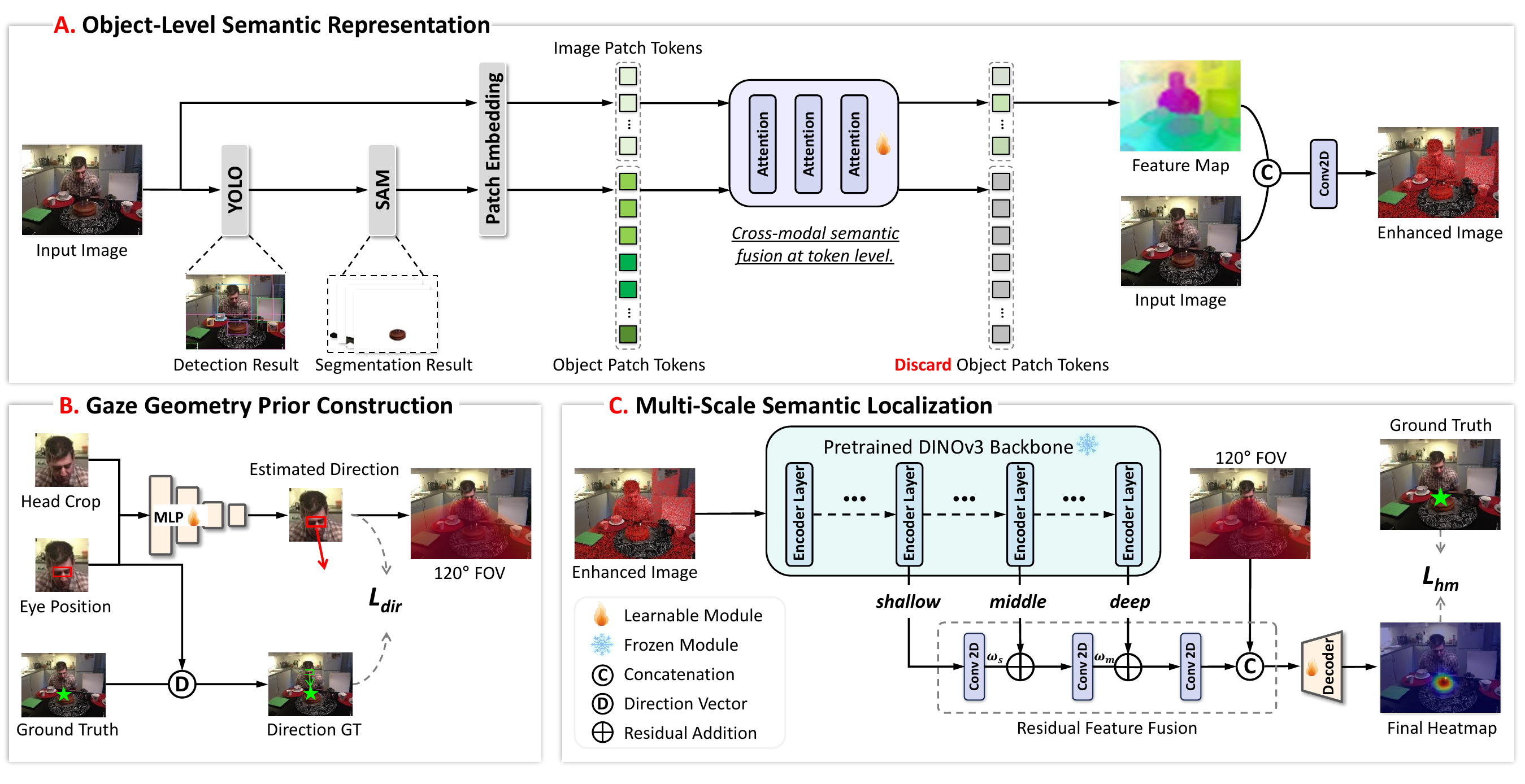}
    \caption{Overview of the proposed framework. (A) Object-level semantic representation introduces object-aware tokens to enhance features through cross-modal fusion. (B) Gaze geometry prior construction estimates gaze direction from head appearance and eye position to generate a FOV prior. (C) Multi-scale semantic localization integrates hierarchical backbone features with the FOV prior to produce the gaze heatmap.}
    \label{fig:overview}
\end{figure}

Building upon these semantic and geometric priors, multi-scale representations are jointly exploited to refine gaze localization within valid candidate regions. As illustrated in Fig.~\ref{fig:overview}, the entire reasoning process is implemented within a unified single-modality framework based on a frozen DINOv3 backbone, where lightweight learnable modules progressively transform object-aware representations into precise gaze predictions with interpretable steps.

\subsection{Object-Level Semantic Representation}

Existing gaze target estimation methods typically predict gaze locations directly from global scene features, lacking explicit modeling of potential gaze objects. As a result, predictions often become unstable in complex scenes containing multiple competing targets or strong background distractions. Considering that human visual attention is inherently object-oriented, we introduce an object-level semantic representation to explicitly construct candidate gaze hypotheses during feature encoding, allowing subsequent reasoning to operate over discrete semantic entities rather than unconstrained spatial responses.

As illustrated in Fig.~\ref{fig:overview}(A), given an input image $\mathbf{I}\in\mathbb{R}^{3\times H\times W}$, we first obtain $N$ candidate object masks $\{\mathbf{M}_i\}_{i=1}^{N}$, where $\mathbf{M}_i\in\{0,1\}^{H\times W}$, using pretrained object detection and segmentation models. The image is divided into non-overlapping patches of size $p\times p$, which are projected into image tokens:
\begin{equation}
	\mathbf{T}_{\text{img}}\in\mathbb{R}^{N_p\times d},
	\quad
	N_p=\frac{H}{p}\cdot\frac{W}{p},
\end{equation}
where $d$ denotes the token embedding dimension.

For each object region, masked image features are encoded into an object token sequence:
\begin{equation}
	\mathbf{T}^{(i)}_{\text{obj}}\in\mathbb{R}^{N_o\times d},
\end{equation}
where $N_o$ denotes the number of tokens used to represent each object, obtained through fixed-grid pooling with $N_o=n^2$. All object tokens are concatenated as:
\begin{equation}
	\mathbf{T}_{\text{obj}}
	=
	[\mathbf{T}^{(1)}_{\text{obj}};
	\ldots;
	\mathbf{T}^{(N)}_{\text{obj}}]
	\in
	\mathbb{R}^{(N\cdot N_o)\times d}.
\end{equation}

The image tokens and object tokens are combined into a unified sequence:
\begin{equation}
	\mathbf{T}_{\text{fuse}}
	=
	[\mathbf{T}_{\text{img}};
	\mathbf{T}_{\text{obj}}]
	\in
	\mathbb{R}^{(N_p+N\cdot N_o)\times d},
\end{equation}
which is processed by a Transformer encoder to enable cross-token interaction. Through this interaction, semantic information associated with individual objects is propagated into the global visual representation, transforming pixel-based scene features into object-aware semantic representations and explicitly establishing candidate gaze hypotheses.

Finally, an object-aware spatial response
$\mathbf{O}\in\mathbb{R}^{1\times H\times W}$
is reconstructed from the fused representation and integrated with the original image through a lightweight fusion mapping:
\begin{equation}
	\mathbf{I}_{\text{enh}}
	=
	\mathcal{F}(\mathbf{I},\mathbf{O}),
\end{equation}
where $\mathcal{F}(\cdot)$ denotes a feature fusion operation realized through channel concatenation followed by a $1\times1$ convolution. The enhanced representation $\mathbf{I}_{\text{enh}}$ preserves original appearance information while incorporating structured object-level semantics, providing a semantic foundation for subsequent geometry-guided reasoning and multi-scale gaze localization.

\subsection{Gaze Geometry Prior Construction}
\label{sec:ggpc}

After obtaining object-level semantic representations, scene semantics alone remain insufficient to uniquely determine gaze locations, since human gaze behavior is inherently constrained by the geometric state of the observer. In practice, visual attention is limited by head orientation and gaze direction, restricting attention to a finite forward-facing visual field. To incorporate such constraints, we explicitly estimate the observer's gaze direction and construct a geometry-aware spatial prior that limits the gaze inference space.

Given the head region image and the corresponding eye position coordinates $(x_e, y_e)$, we first extract head appearance features $\mathbf{f}_{\text{head}}$. The head feature and eye position are jointly fed into a multilayer perceptron to predict a normalized two-dimensional gaze direction vector:
\begin{equation}
	\hat{\mathbf g}
	=
	\frac{
		\mathrm{MLP}_{dir}
		([\mathbf f_{\text{head}}, x_e, y_e])
	}{
		\left\|
		\mathrm{MLP}_{dir}
		([\mathbf f_{\text{head}}, x_e, y_e])
		\right\|
	}.
\end{equation}
The predicted vector represents the estimated gaze direction of the observer, as illustrated in Fig.~\ref{fig:overview}(B).

To stabilize direction learning, a supervision signal is constructed from the ground-truth target. Let $\mathbf p_e$ denote the eye position and $\mathbf p_t$ denote the ground-truth gaze target position. The ground-truth gaze direction is defined as:
\begin{equation}
	\mathbf g^{gt}
	=
	\frac{\mathbf p_t-\mathbf p_e}
	{\|\mathbf p_t-\mathbf p_e\|}.
\end{equation}
The predicted direction is constrained using a directional loss:
\begin{equation}
	\mathcal{L}_{\text{dir}}
	=
	1-\hat{\mathbf g}\cdot\mathbf g^{gt}.
\end{equation}

Based on the estimated gaze direction, a field-of-view (FOV) geometric prior is constructed. For any pixel location $(x,y)$, let $\hat{\mathbf v}_{(x,y)}$ denote the unit vector from the eye center. The directional response is computed via cosine similarity:
\begin{equation}
	c_{(x,y)}
	=
	\hat{\mathbf v}_{(x,y)}\cdot\hat{\mathbf g}.
\end{equation}
Considering the bounded nature of human vision, responses are thresholded by the FOV range $\theta$, where $\theta = 120^\circ$ (Table~\ref{tab:range}), to obtain the cone map:
\begin{equation}
	C_{(x,y)}=
	\begin{cases}
		c_{(x,y)}, & c_{(x,y)}>\cos\theta,\\
		0, & \text{otherwise}.
	\end{cases}
\end{equation}

To mitigate discontinuities near the cone boundary, the cone response is smoothed using Gaussian filtering, yielding FOV representation $C_{\text{smooth}}$. A learnable FOV embedding vector $\mathbf e_{fov}\in\mathbb{R}^{d'}$ is then broadcast-multiplied with the smoothed cone map to produce a geometry-aware representation:
\begin{equation}
	\mathbf E_{fov}
	=
	C_{\text{smooth}}\odot \mathbf e_{fov}.
\end{equation}

The resulting feature map $\mathbf E_{fov}$ explicitly encodes spatial regions that are visually reachable by the observer and serves as the gaze geometry prior injected into the subsequent multi-scale localization stage, guiding gaze prediction within geometrically feasible regions.

\subsection{Multi-Scale Semantic Localization}

After establishing object-level semantic hypotheses and constraining visually reachable regions through gaze geometry priors, precise gaze prediction requires fine-grained localization within semantically consistent and geometrically feasible regions. As illustrated in Fig.~\ref{fig:overview}(C), we jointly integrate semantic representations and observer-dependent geometric constraints within a multi-scale feature space, enabling geometry-guided gaze localization and completing the hierarchical reasoning process from object selection to spatial refinement.

The enhanced representation $\mathbf I_{\text{enh}}$ is fed into a frozen DINOv3 backbone to extract hierarchical visual features from layers. Features at different depths encode complementary information, where shallow layers preserve spatial structures while deep layers capture high-level semantics. All feature maps are projected into a shared embedding space to facilitate cross-scale interaction.

Instead of directly aggregating multi-scale features, we treat deep semantic features as the dominant representation and introduce intermediate and shallow features as bounded residual complements:
\begin{equation}
	\mathbf F_{\text{fused}}
	=
	\mathbf F_{\text{deep}}
	+
	\alpha_{\text{mid}}\mathbf F_{\text{mid}}
	+
	\alpha_{\text{shallow}}\mathbf F_{\text{shallow}},
\label{eq:fuse}
\end{equation}
where $\alpha_{\text{mid}}$ and $\alpha_{\text{shallow}}$ denote multi-scale fusion coefficients, with hyperparameter analysis reported in Table~\ref{tab:alpha}.

The gaze geometry priors constructed in Sec.~\ref{sec:ggpc} are subsequently injected into the fused representation. Specifically, the field-of-view feature $\mathbf E_{fov}$, encoding geometrically reachable regions along the estimated gaze direction, is fused with semantic representations, allowing gaze prediction to be simultaneously guided by object semantics and gaze-reachable spatial constraints:
\begin{equation}
	\mathbf F_{\text{guided}}
	=
	\mathrm{Fusion}
	(\mathbf F_{\text{fused}}, \mathbf E_{fov}).
\end{equation}

Finally, the geometry-guided representation is processed by a lightweight decoding network to progressively recover spatial resolution and produce the final gaze probability heatmap $\mathbf H$. The resulting prediction no longer relies on global pixel-wise regression, but instead emerges from the joint interaction between object-level semantic hypotheses and observer-dependent geometric constraints, enabling structured gaze localization.

\subsection{Training Objective}
The proposed framework is optimized under joint supervision for gaze localization and gaze existence prediction, enabling consistent learning of spatial reasoning under semantic and geometric constraints.

For gaze localization, the ground-truth fixation point is represented as a two-dimensional Gaussian heatmap $\mathbf H^{gt}$ centered at the annotated gaze target. The predicted heatmap $\mathbf H$ is optimized using pixel-wise binary cross-entropy loss:
\begin{equation}
	\mathcal L_{\text{hm}}
	=
	-\sum_{x,y}
	\Big(
	\mathbf H^{gt}_{x,y}\log \mathbf H_{x,y}
	+
	(1-\mathbf H^{gt}_{x,y})
	\log(1-\mathbf H_{x,y})
	\Big).
\end{equation}

To handle cases where the gaze target lies outside the image, an additional in/out prediction branch is introduced. Let $p$ denote the predicted probability indicating whether the gaze target is inside the image and $y\in\{0,1\}$ the corresponding ground-truth label. The in/out classification loss is defined as:
\begin{equation}
	\mathcal L_{\text{inout}}
	=
	-y\log p-(1-y)\log(1-p).
\end{equation}

The overall training objective is formulated as:
\begin{equation}
	\mathcal L
	=
	\mathcal L_{\text{hm}}
	+
	\lambda_{\text{dir}}\mathcal L_{\text{dir}}
	+
	\lambda_{\text{inout}}\mathcal L_{\text{inout}},
\label{eq:loss}
\end{equation}
where $\mathcal{L}_{\text{dir}}$ denotes the direction supervision defined in Sec.~\ref{sec:ggpc}, and $\lambda_{\text{dir}}$ is detailed in Table~\ref{tab:lamda}. 
$\lambda_{\text{inout}} \in \{0,1\}$ is a dataset-dependent binary hyperparameter, set to 1 when in/out-of-frame annotations are available and 0 otherwise. 
The weighting coefficients balance supervision signals during optimization.

\section{Experiments}

\subsection{Experimental Setup, Datasets, and Metrics}
\noindent\textbf{Implementation Details.}
All experiments are conducted on a single NVIDIA RTX 5090 GPU.
A pretrained DINOv3 ViT-L/16 backbone is employed and kept entirely frozen during training.
Object masks are generated offline using YOLO11x and SAM2-hiera-large\cite{ravi2024sam}.
Input images are resized to $512\times512$, and gaze heatmaps are predicted at $64\times64$ resolution.
Training is performed using Adam\cite{diederik2014adam} with an initial learning rate of $1\times10^{-3}$.
The overall training objective jointly optimizes heatmap regression, direction supervision, and in/out classification losses. More details are provided in the \textit{supplementary material}.

\noindent\textbf{Datasets.}
Experiments are conducted on four public gaze estimation benchmarks: GazeFollow\cite{recasens2015they}, VideoAttentionTarget\cite{chong2020detecting}, ChildPlay\cite{tafasca2023childplay}, and GOO-Real\cite{tomas2021goo}.
GazeFollow provides image-based gaze annotations with multiple observers per sample, enabling robust evaluation under gaze ambiguity.
VideoAttentionTarget extends the task to video sequences and includes gaze in/out-of-frame annotations.
ChildPlay evaluates gaze prediction in child-interaction scenarios characterized by larger behavioral variability.
GOO-Real focuses on real-world scenarios with notable domain shifts, serving as a benchmark for cross-domain generalization and distribution shift robustness.

\noindent\textbf{Evaluation Metrics.}
Performance is evaluated using standard metrics including AUC and L2 distance under official evaluation protocols for fair and consistent comparison.
For GazeFollow, both Avg L2 and Min L2 are reported due to multiple gaze annotations per image.
Average Precision (AP) is additionally adopted for datasets involving gaze in/out-of-frame prediction tasks.
\begin{table}[!t]
	\centering
	\caption{Quantitative comparisons on (A) GazeFollow, (B) VideoAttentionTarget, (C) ChildPlay, and (D) GOO-Real. Higher AUC/AP and lower L2 indicate better performance. Bold and underline denote best and second-best.$^\ast$ estimated parameters.}
	\begin{tabular}{c}
		\includegraphics[width=1\linewidth]{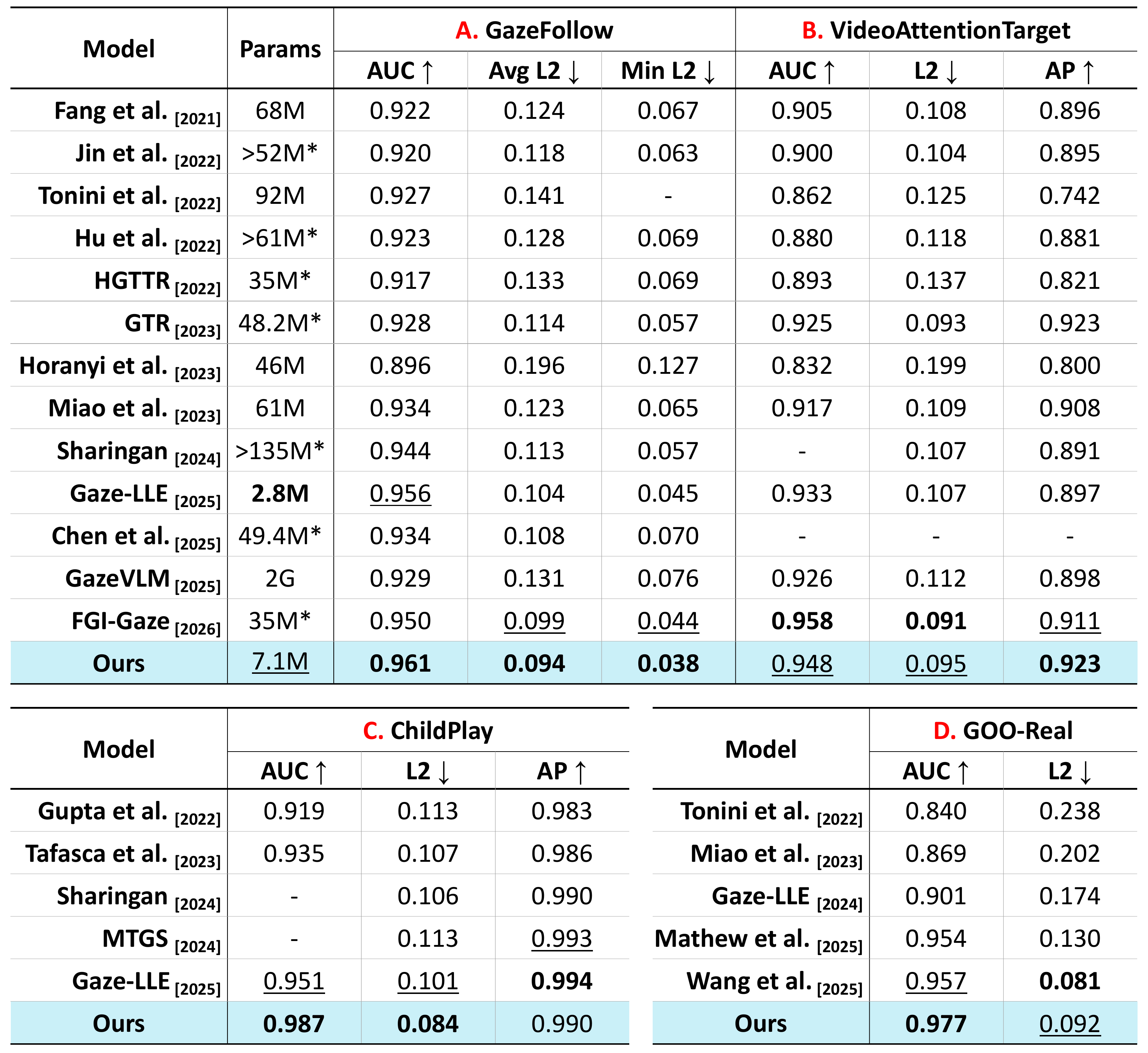}
	\end{tabular}
	\label{tab:tab1}
\end{table}

\subsection{Comparison with Existing Methods}
We comprehensively compare our method with a broad range of existing gaze target estimation approaches across all evaluated benchmarks.
On GazeFollow and VideoAttentionTarget, comparisons include Fang et al.\cite{lian2018believe}, Jin et al.\cite{jin2022depth}, Tonini et al.\cite{tonini2022multimodal}, Hu et al.\cite{hu2022gaze}, HGTTR\cite{tu2022end}, GTR\cite{tu2023gaze}, Horanyi et al.\cite{horanyi2023they}, Miao et al.\cite{miao2023patch}, Sharingan\cite{tafasca2024sharingan}, Gaze-LLE\cite{ryan2025gaze}, Chen et al.\cite{chen2025privileged}, GazeVLM\cite{mathew2025gazevlm}, and FGI-Gaze\cite{lan2026fgi}.
On ChildPlay, we further compare with Gupta et al.\cite{gupta2022modular}, Tafasca et al.\cite{tafasca2023childplay}, Sharingan\cite{tafasca2024sharingan}, MTGS\cite{gupta2024mtgs}, and Gaze-LLE\cite{ryan2025gaze}.
For the cross-domain benchmark GOO-Real, comparisons involve Tonini et al.\cite{tonini2022multimodal}, Miao et al.\cite{miao2023patch}, Gaze-LLE\cite{ryan2025gaze}, Mathew et al.\cite{mathew2025leveraging}, and Wang et al.\cite{wang2024transgop}, covering representative methods under diverse evaluation settings and experimental conditions.

\textbf{Quantitative Comparisons.}
As shown in Table~\ref{tab:tab1}, the proposed method achieves consistently strong performance across all evaluated benchmarks. On GazeFollow, our framework obtains the highest AUC together with lower Avg L2 and Min L2 errors, indicating improved spatial consistency and more precise gaze localization. On VideoAttentionTarget, the proposed method maintains competitive performance in both localization accuracy and in/out-of-frame prediction, demonstrating robustness under dynamic scene variations without explicit temporal modeling. For ChildPlay, which involves complex interaction behaviors and increased appearance diversity, our approach achieves clear improvements in both AUC and L2 metrics, suggesting enhanced capability in resolving competing attention targets. On the cross-domain benchmark GOO-Real, our method achieves the best AUC with only lightweight fine-tuning, highlighting strong generalization ability to real-world scenarios. Notably, these results are achieved with only 7.1M parameters, providing a favorable balance between prediction accuracy and model efficiency compared with existing approaches.

\begin{figure}[!t]
	\centering
	\includegraphics[width=\textwidth]{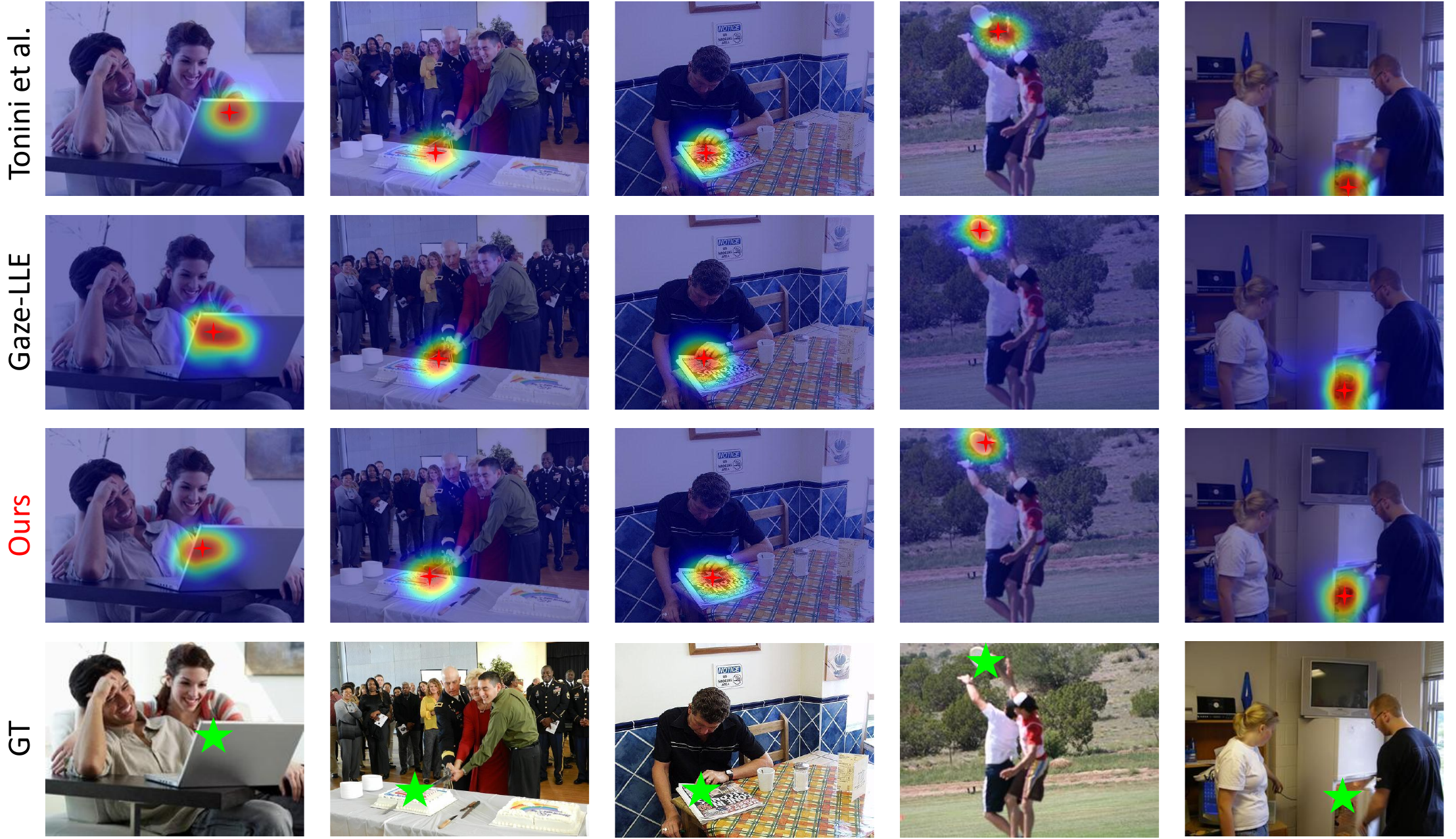}
	\caption{Qualitative comparison of gaze prediction results on representative examples.}
	\label{fig:comp}
\end{figure}

\textbf{Qualitative Analysis.}
The qualitative results shown in Fig.~\ref{fig:comp} further validate the effectiveness of the proposed method in complex real-world scenarios. Compared with existing approaches such as Tonini et al.\cite{tonini2023object} and Gaze-LLE\cite{ryan2025gaze}, which often produce scattered or shifted responses under multiple candidate targets or background distractions, our method generates more compact predictions that closely align with the true gaze targets. This advantage arises from reformulating gaze target estimation as a hierarchical reasoning process, where object-level semantic modeling identifies potential attention entities while geometrically reachable regions constrain the spatial search space. Such a design effectively suppresses irrelevant distractions and alleviates attention ambiguity. As a result, the predicted heatmaps exhibit clearer spatial concentration and stronger object consistency, indicating that the model performs stable structured gaze decision making rather than relying on pixel-level regression. These visualization results are consistent with quantitative findings, further demonstrating the robustness and reliability of the proposed framework in complex interaction and real-world environments.

\subsection{Ablation Study}
\textbf{Component Ablation.}
Table~\ref{tab:component} presents ablation results by progressively enabling object-level fusion, multi-scale representation, and gaze geometry prior modules (FOV).
Starting from the baseline model without additional components (No.~0), introducing object-level semantic fusion improves performance on both datasets, indicating that explicit object-aware representations reduce attention ambiguity. Incorporating multi-scale feature modeling further improves localization accuracy, demonstrating the importance of leveraging hierarchical visual representations beyond single-layer features.
When the gaze FOV prior is introduced, performance improves by constraining predictions within geometrically feasible regions. Combining semantic modeling with geometric guidance produces more stable spatial responses than using either component alone.

The complete model integrating all three components achieves the best performance across all evaluation metrics, confirming that object-level reasoning, multi-scale representation, and geometry-aware priors contribute complementarily to accurate gaze target estimation on tested benchmarks.

\begin{table}[!t]
	\centering
	\caption{Ablation study on key components across GazeFollow and VideoAttentionTarget, with bold entries denoting the best results.}
	\begin{tabular}{c}
		\includegraphics[width=1\linewidth]{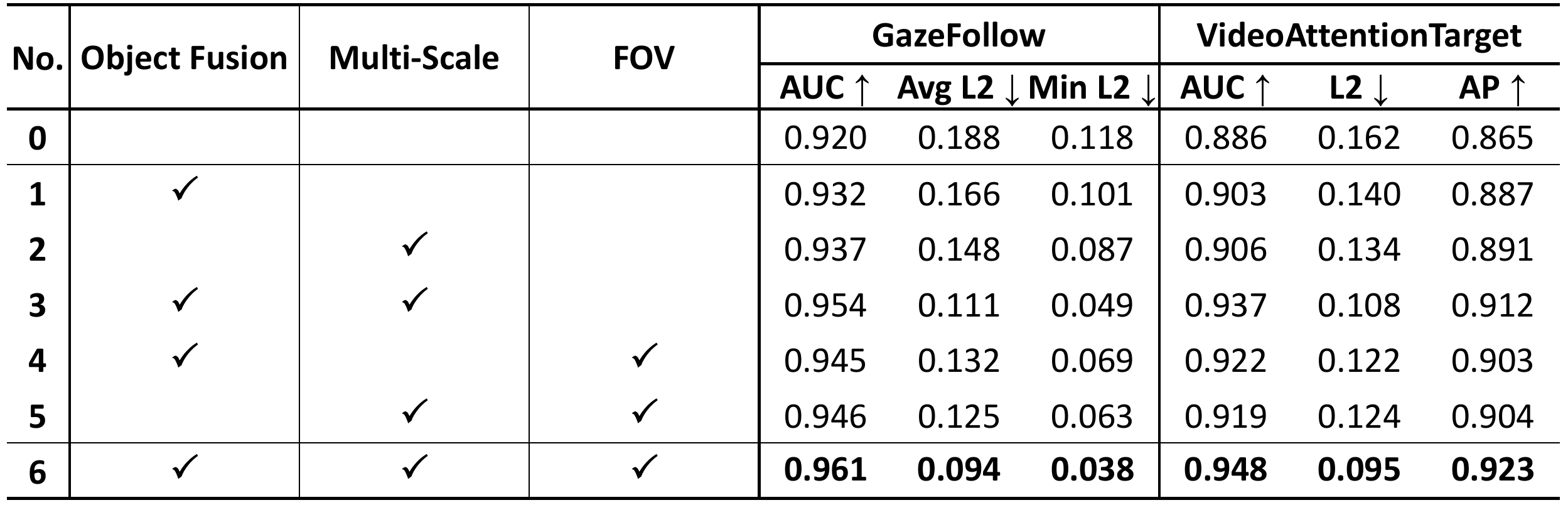}
	\end{tabular}
	\label{tab:component}
\end{table}

\begin{table}[t]
	\centering
	
	\begin{minipage}{0.48\textwidth}
		\centering
		\caption{Ablation study on the impact of multi-scale feature levels.}
		\begin{tabular}{c}
		\includegraphics[width=1\linewidth]{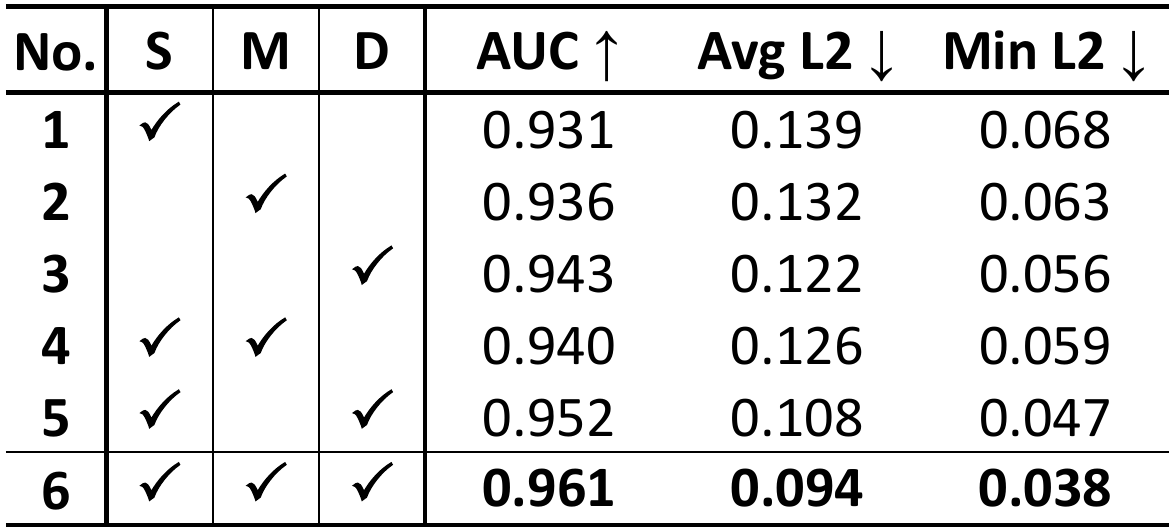}
		\end{tabular}
		\label{tab:smd}
	\end{minipage}
	\hfill
	\begin{minipage}{0.48\textwidth}
		\centering
		\caption{Comparison of different backbone architectures.}
		\begin{tabular}{c}
			\includegraphics[width=1\linewidth]{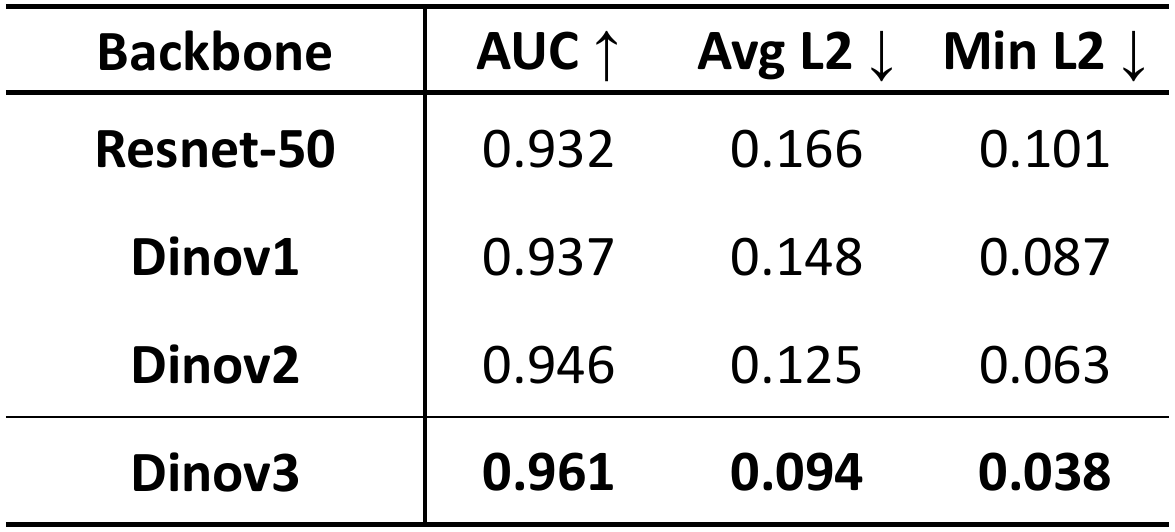}
		\end{tabular}
		\label{tab:backbone}
	\end{minipage}
\end{table}

\textbf{Multi-scale Ablation.}
Table~\ref{tab:smd} studies the impact of selecting backbone features from different depth levels for multi-scale semantic localization, where S, M, and D denote shallow, middle, and deep features, respectively. Using a single feature level provides reasonable performance, while deeper representations yield stronger semantics and lower localization error. Performance improves progressively as features from multiple depth levels are incorporated, indicating that spatial details from shallow layers and semantic cues from deeper layers are complementary. The best results are achieved when all feature levels are utilized, validating the effectiveness of the proposed multi-scale fusion strategy.

\textbf{Backbone Analysis.}
Table~\ref{tab:backbone} compares different backbone architectures for feature extraction. Performance improves consistently from ResNet-50\cite{he2016deep} to DINO-based models\cite{caron2021emerging,oquab2023dinov2,simeoni2025dinov3}, highlighting the advantage of pretrained vision foundation models in capturing semantic representations relevant to gaze understanding. The DINOv3 backbone achieves the best results across all metrics, suggesting that stronger hierarchical representations further significantly enhance the effectiveness of the proposed reasoning framework.

\begin{table}[t]
	\centering
	
	\begin{minipage}{0.48\textwidth}
		\centering
		\caption{Ablation study on attention layer numbers in cross-modal fusion.}
		\begin{tabular}{c}
			\includegraphics[width=1\linewidth]{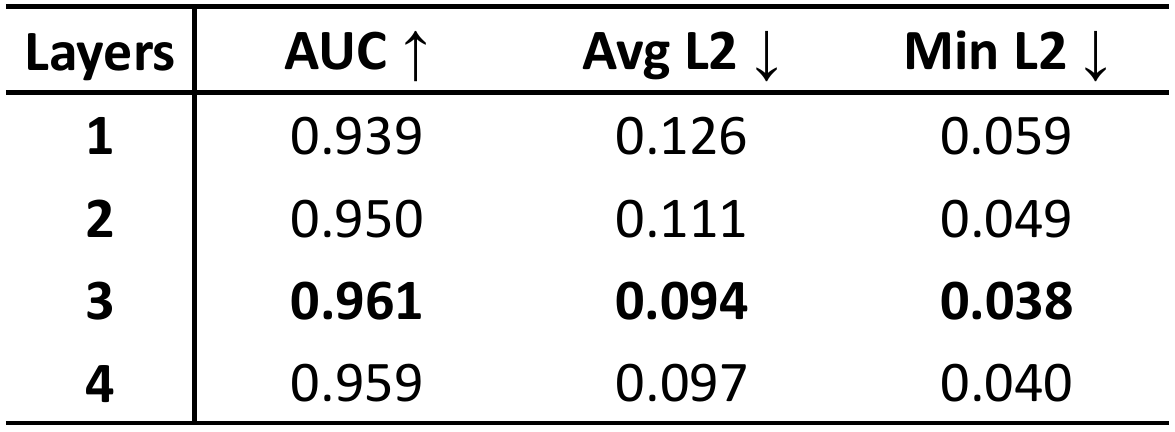}
		\end{tabular}
		\label{tab:layers}
	\end{minipage}
	\hfill
	\begin{minipage}{0.48\textwidth}
		\centering
		\caption{Impact of different FOV ranges on gaze estimation performance.}
		\begin{tabular}{c}
			\includegraphics[width=1\linewidth]{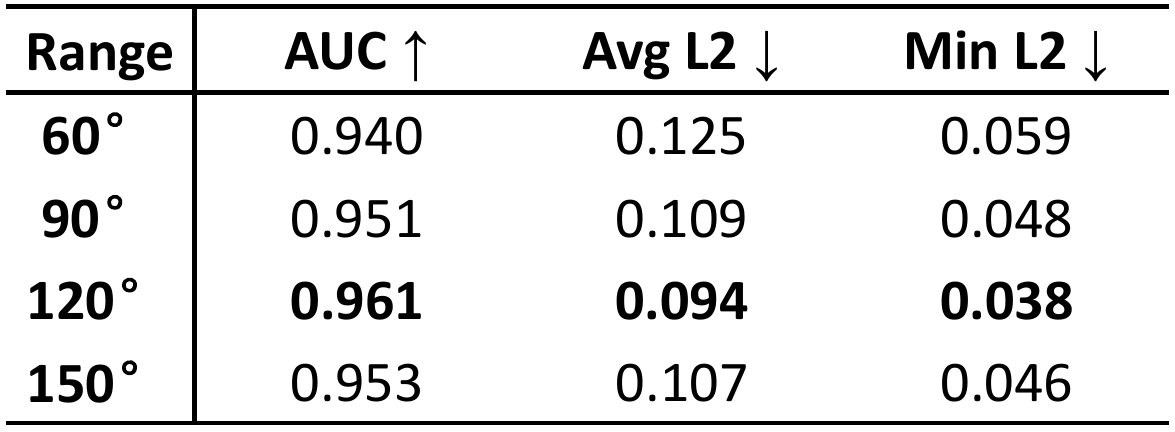}
		\end{tabular}
		\label{tab:range}
	\end{minipage}
\end{table}

\begin{table}[t]
	\centering
	
	\begin{minipage}{0.48\textwidth}
		\centering
		\caption{Ablation study on the residual fusion weights defined in Eq.~\ref{eq:fuse}.}
		\begin{tabular}{c}
			\includegraphics[width=1\linewidth]{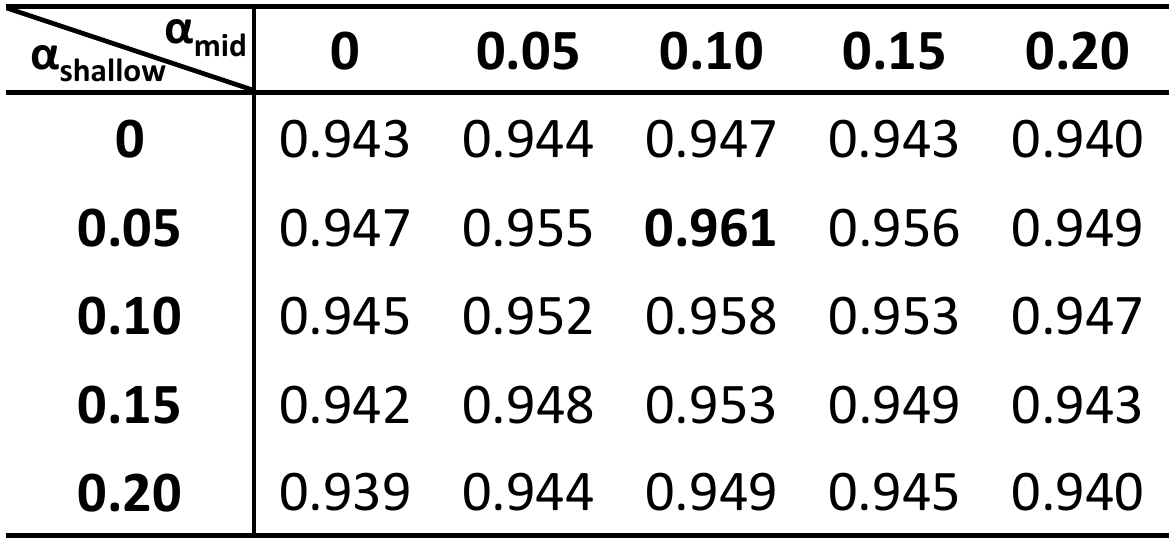}
		\end{tabular}
		\label{tab:alpha}
	\end{minipage}
	\hfill
	\begin{minipage}{0.48\textwidth}
		\centering
		\caption{Ablation study on the direction supervision coefficient $\lambda_{\text{dir}}$ in Eq.~\ref{eq:loss}.}
		\begin{tabular}{c}
			\includegraphics[width=1\linewidth]{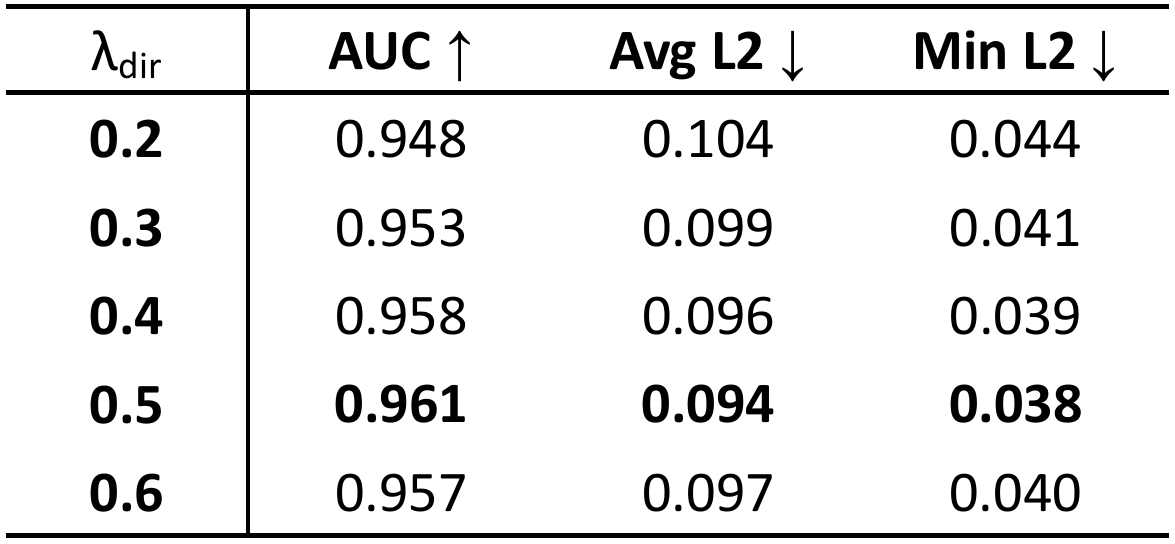}
		\end{tabular}
		\label{tab:lamda}
	\end{minipage}
\end{table}

\subsection{Parameter Sensitivity Analysis}

To further analyze the robustness of the proposed framework, we conduct parameter sensitivity experiments on key components in semantic fusion, geometric prior modeling, multi-scale feature integration, and training optimization.

\textbf{Effect of Attention Layer Depth.}
Table~\ref{tab:layers} evaluates the influence of the number of attention layers in the cross-modal semantic fusion module. Increasing attention depth enhances object-level semantic interaction and improves gaze localization accuracy. Performance peaks when three attention layers are adopted, indicating that sufficient semantic reasoning is achieved at moderate depth. Further increasing the layer number slightly degrades performance, which may be attributed to feature over-smoothing and redundant token interaction.

\textbf{Effect of FOV Range.}
Table~\ref{tab:range} studies the impact of different field-of-view (FOV) ranges in the geometry prior construction stage. A narrow FOV overly restricts the search space, while an excessively large FOV weakens geometric guidance. The best performance is achieved with a $120^\circ$ FOV, suggesting that a balanced geometric constraint effectively limits attention to visually reachable regions without suppressing valid gaze targets. Visualizations of the FOV cone map are provided in the \textit{supplementary material}.

\textbf{Effect of Multi-scale Residual Fusion Weights.}
Table~\ref{tab:alpha} analyzes the residual fusion coefficients defined in Eq.~\ref{eq:fuse}, where deep semantic features serve as the primary representation while shallow and middle features are introduced as residual refinements controlled by $\alpha_{\text{shallow}}$ and $\alpha_{\text{mid}}$. The results show that moderate residual contributions improve localization accuracy, whereas excessive weighting introduces feature interference and slightly degrades performance.

\textbf{Effect of Direction Supervision Weight.}
Table~\ref{tab:lamda} investigates the influence of the direction supervision coefficient $\lambda_{\text{dir}}$ in the overall training objective (Eq.~\ref{eq:loss}). Increasing the weight initially improves performance by strengthening geometric consistency during optimization. However, excessive weighting slightly harms localization accuracy due to over-constraining the learning process. The optimal performance is obtained at $\lambda_{\text{dir}}=0.5$, demonstrating an effective balance between heatmap supervision and geometric direction learning.

Overall, the proposed framework exhibits stable performance across diverse parameter settings, indicating robustness and balanced interactions among semantic reasoning, geometric constraints, and optimization objectives. Further ablation results are provided in the \textit{supplementary material}.

\begin{figure}[!t]
	\centering
	\includegraphics[width=\textwidth]{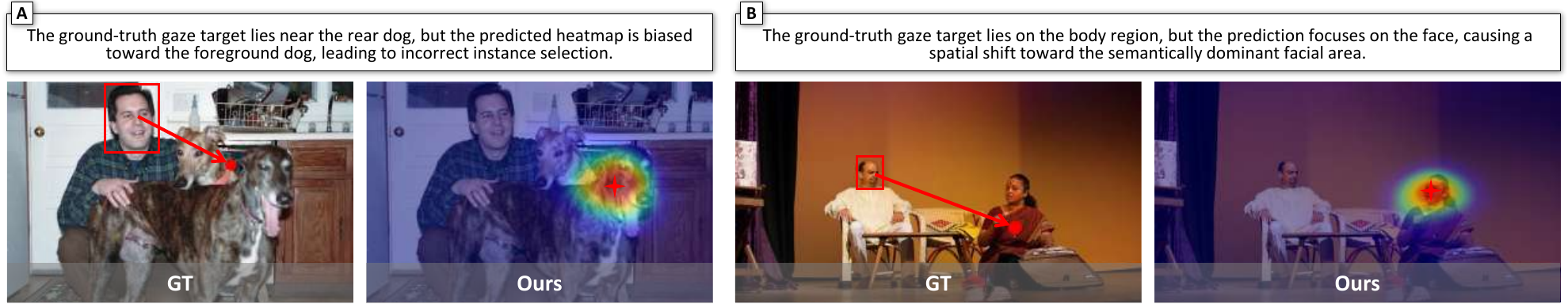}
	\caption{Failure cases of the proposed method.}
	\label{fig:failure}
\end{figure}

\section{Limitation and Future Work}
Despite the strong performance of the proposed framework, several limitations remain, mainly arising under two challenging scenarios.

As illustrated in Fig.~\ref{fig:failure}(A), when multiple semantically similar objects appear in close spatial proximity, visually dominant instances may generate stronger responses despite consistent geometric cues, leading to incorrect target selection. This reflects the difficulty of resolving semantic competition among candidate objects in complex scenes.
As shown in Fig.~\ref{fig:failure}(B), ambiguity commonly occurs in human-centered scenes where gaze targets correspond to extended regions rather than precise points. In such cases, the model tends to focus on semantically salient areas, such as faces, instead of annotated locations, revealing a mismatch between discrete supervision and continuous human attention distribution.

Future work explores adaptive object reasoning mechanisms and uncertainty-aware gaze modeling to better address semantic ambiguity and multi-object competition. Extending the framework to multi-person interaction scenarios and temporal gaze reasoning in videos also represents a promising direction.

\section{Conclusion}

In this work, we reformulate gaze target estimation as a hierarchical reasoning problem instead of direct pixel-level regression. 
The proposed framework performs object-level semantic selection together with geometry-guided spatial localization within a unified single-modality architecture. 
By integrating multi-scale feature fusion and gaze geometry priors, the method achieves stable and interpretable gaze prediction in complex scenes. 
Extensive experiments demonstrate consistent improvements over existing approaches, highlighting the effectiveness of structured semantic and geometric reasoning for gaze estimation.


\medskip
\noindent\textbf{Acknowledgements.} This work was supported in part by the National Natural Science Foundation of China under Grant 62172246, in part by Excellent Young Scientists Fund of Natural Science Foundation of Shandong Province under Grant ZR2024YQ071, in part by the Key Laboratory of Forensic Examination for Sichuan Provincial Universities under Grant 2024YB01, and in part by the Fundamental Research Funds for the Central Universities under Grant 22CX06037A.

%
%
\bibliographystyle{splncs04}
\bibliography{main}
\end{document}